\documentclass[journal=jctc,manuscript=article]{achemso}
\usepackage{chemformula} 
\ProvideTextCommand{\DJ}{OT1}{\leavevmode\raisebox{-.5ex}{\makebox[0pt][l]{\hskip-.07em\accent"16\hss}}D}
\usepackage{adjustbox}
\usepackage{caption}
\usepackage{subcaption}
\usepackage{color}
\usepackage{booktabs}  
\usepackage{tabularx}  
\usepackage{array}     
\usepackage{ragged2e}  
\usepackage{multirow}  
\usepackage{hyperref}
\usepackage{booktabs}
\usepackage{multirow}

\usepackage[flushleft]{threeparttable}
\usepackage{booktabs,caption}
\usepackage{amsmath}
\usepackage{makecell}
\usepackage{tabularx}
\usepackage{graphicx}   
\usepackage{amsmath}    
\usepackage{amssymb}    
\usepackage{textcomp}   
\usepackage{upgreek}   
\usepackage{chemmacros} 
\usepackage{svg}

\usepackage{booktabs}
\usepackage{tabularx}
\usepackage{array}
\usepackage{ragged2e}
\usepackage{xcolor}
\usepackage{colortbl}

\definecolor{headerblue}{RGB}{52, 73, 94}
\definecolor{lightgray}{RGB}{245, 245, 245}
\author{Zekun Jiang}
\affiliation[CUPB]{State Key Laboratory of Heavy Oil Processing, College of Carbon Neutrality Future Technology, China University of Petroleum (Beijing), Beijing 102249, China}

\author{Chunming Xu}
\affiliation[CUPB]{State Key Laboratory of Heavy Oil Processing, College of Carbon Neutrality Future Technology, China University of Petroleum (Beijing), Beijing 102249, China}

\author{Tianhang Zhou}
\affiliation[CUPB]{State Key Laboratory of Heavy Oil Processing, College of Carbon Neutrality Future Technology, China University of Petroleum (Beijing), Beijing 102249, China}
\email{zhouth@cup.edu.cn}

\title[An \textsf{achemso} demo]
  {Cyber Academia-Chemical Engineering (CA-ChemE): A Living Digital Town for Self-Directed Research Evolution and Emergent Scientific Discovery}

\abbreviations{IR,NMR,UV}

\DeclareUnicodeCharacter{2212}{-}

\begin{document}
\pagebreak

\begin{abstract}
The rapid advancement of artificial intelligence (AI) has demonstrated substantial potential in chemical engineering, yet existing AI systems remain limited in interdisciplinary collaboration and exploration of uncharted problems. To address these issues, we present the Cyber Academia-Chemical Engineering (CA-ChemE) system, a living digital town that enables self-directed research evolution and emergent scientific discovery through multi-agent collaboration. By integrating domain-specific knowledge bases, knowledge enhancement technologies, and collaboration agents, the system successfully constructs an intelligent ecosystem capable of deep professional reasoning and efficient interdisciplinary collaboration. Our findings demonstrate that knowledge base-enabled enhancement mechanisms improved dialogue quality scores by 10--15\% on average across all seven expert agents, fundamentally ensuring technical judgments are grounded in verifiable scientific evidence. However, we observed a critical bottleneck in cross-domain collaboration efficiency, prompting the introduction of a Collaboration Agent (CA) equipped with ontology engineering capabilities. CA's intervention achieved 8.5\% improvements for distant-domain expert pairs compared to only 0.8\% for domain-proximate pairs---a 10.6-fold difference---unveiling the ``diminished collaborative efficiency caused by knowledge-base gaps'' effect. This study demonstrates how carefully designed multi-agent architectures can provide a viable pathway toward autonomous scientific discovery in chemical engineering.
\end{abstract}

\pagebreak



\maketitle
\section{Introduction}\label{sec1}

The rapid advancement of artificial intelligence (AI) technologies has demonstrated substantial potential in chemical engineering, with deep learning and generative model-based approaches achieving significant progress across specialized professional domains. At the molecular level, machine learning models including variational autoencoders, generative adversarial networks, and transformer-based architectures have exhibited exceptional performance in molecular property prediction, materials design and discovery, successfully generating novel molecular structures with desired properties \cite{molecular_design_jacs2021, drug_discovery_nature2022, transformer_molecules2023}. At the materials level, AI-driven high-throughput computational screening has markedly accelerated the discovery of novel catalysts, battery materials, and photovoltaic compounds \cite{materials_discovery_science2021, catalyst_design_aiche2022, battery_materials_nature2023}. At the system level, AI technologies have achieved comprehensive coverage spanning from quantum-level molecular interactions to plant-scale process optimization, demonstrating superior application outcomes in critical areas including reaction mechanism elucidation, process optimization, computational fluid dynamics simulations, safety assessment, and sustainability evaluation \cite{multiscale2025, review_mdpi2022, process_control_cep2022, safety_ai_loss2023, chem_ai_nature2021}. Concurrently, the successful integration of physics-informed neural networks (PINNs) and hybrid modeling approaches has further enhanced the capability of solving complex partial differential equations in chemical processes, substantially improving the prediction accuracy of transport phenomena and reaction kinetics \cite{pinns_chemical2022, hybrid_modeling_computers2023}. Furthermore, the combination of multi-agent systems with domain-specific knowledge graphs has exhibited considerable potential in automated knowledge extraction, semantic alignment, complex task allocation, and collaborative problem-solving, thereby establishing foundations for constructing intelligent scientific and engineering agent ecosystems \cite{scieng_sciagents2024, openreview_kg_agents, knowledge_graphs_chem2023}.

Despite remarkable achievements of artificial intelligence across individual specialized domains in chemical engineering, existing systems continue to encounter fundamental limitations in interdisciplinary collaboration and knowledge integration, particularly manifesting severe inadequacies in automation capabilities. Current AI models predominantly employ supervised learning paradigms, relying on human-annotated data and explicitly defined task objectives to optimize performance within specific professional directions. This passive response mode intrinsically \emph{constrains the developmental boundaries of AI}: systems can only seek solutions within the problem space already known to humans, struggling to explore novel problems that humans have not yet clearly defined or to discover unexpected cross-domain connections \cite{autonomous_agents_challenges2023, human_ai_collaboration2022}. This raises a critical question: \emph{is it feasible to construct AI systems capable of autonomously exploring chemical engineering problems to transcend these boundaries?} Autonomous exploration might enable AI systems to discover hidden connections among molecules, reactions, and processes that humans have not yet recognized, to propose innovative hypotheses that transcend current paradigms, and to generate breakthrough insights at the intersections of multiple specialized domains. However, autonomous exploration by individual AI models confronts formidable challenges. Chemical engineering problems are inherently interdisciplinary in nature; single AI systems lack the comprehensive knowledge coverage requisite for traversing these professional boundaries and cannot establish effective semantic alignment among the disparate terminology systems, knowledge representations, and decision logics across different specialized domains \cite{nsr_deeplearn2024, knowledge_sharing_agents2023}. These isolated systems often demonstrate satisfactory performance in controlled environments, yet prove inadequate when confronting complex interdisciplinary scenarios that necessitate genuine collaborative decision-making \cite{multiscale2025, industrial_ai_gap2023, scale_up_challenges2022}. More critically, existing systems exhibit pronounced semantic gaps and knowledge silos in cross-domain communication, severely restricting both the depth and breadth of autonomous exploration.

In recent years, the AI Town generative agent framework has demonstrated transformative potential for large language model-driven multi-agent interactions within controlled sandbox environments, successfully simulating complex social dynamics \cite{park2023generativeagentsinteractivesimulacra}. Nevertheless, this framework primarily relies on general-purpose pre-trained models which, while possessing extensive commonsense knowledge, lack sufficient domain-specific knowledge depth and specialized reasoning capabilities when confronting sophisticated scientific problems in professional fields such as chemical engineering. To address this limitation and realize the vision of a living digital community characterized by self-directed research evolution, we propose the innovative concept of ``Cyber Academia'' specifically tailored and optimized for chemical engineering scenarios. The fundamental innovation of Cyber Academia resides in two critical breakthroughs: First, each specialized expert model is systematically equipped with a meticulously curated domain-specific knowledge base, achieving a paradigmatic shift from static pre-trained knowledge to dynamic knowledge injection through techniques including retrieval-augmented generation (RAG), domain-adaptive fine-tuning (LoRA), and knowledge graph embeddings \cite{rag_chemical2023, domain_tuning_materials2022, physics_informed_chem2023}. This knowledge enhancement mechanism substantively elevates the professional depth, technical accuracy, and practical operability of each role-specific model, thereby enabling agents to perform reasoning and decision-making based on the latest literature analogous to genuine researchers, supporting self-directed research evolution. Second, recognizing that professional knowledge bases alone prove insufficient for achieving efficient collaboration---as agents from different professional domains exhibit significant disparities in terminology systems, knowledge representation, and decision logic, creating semantic gaps that severely impede cross-domain communication and collaborative decision-making---we strategically introduce a collaboration agent endowed with specialized expertise in ontology engineering. Through ontology engineering practices, this agent assumes responsibility for terminology standardization, context translation, knowledge integration, and strategic decision coordination among diverse expert models, thereby significantly enhancing the efficiency of interdisciplinary collaboration and the practical executability of derived conclusions \cite{ontology_chemical2022, coordination_agents2023}. Through the ontology engineering practices of the collaboration agent, the system can dismantle knowledge silos, facilitate conceptual mapping and cross-domain intelligent emergence, transforming isolated professional perspectives into complementary collaborative resources. Based on this comprehensive framework, we contend that the synergistic combination of knowledge base-driven expert models with collaboration-oriented ontology engineering agents represents the most promising future direction for large language model applications in the chemical engineering domain \cite{park2023generative, scieng_sciagents2024, multiscale2025, multi_agent_nature2023}, providing a viable and scalable technological pathway toward achieving genuine interdisciplinary collaboration and emergent scientific discovery \cite{industry_research_integration2023}.

\section{Methods}
\subsection{Overview of the Cyber Academia Architecture and Expert Systems}
\begin{figure}[H]
\centering
\includegraphics[width=0.9\textwidth]{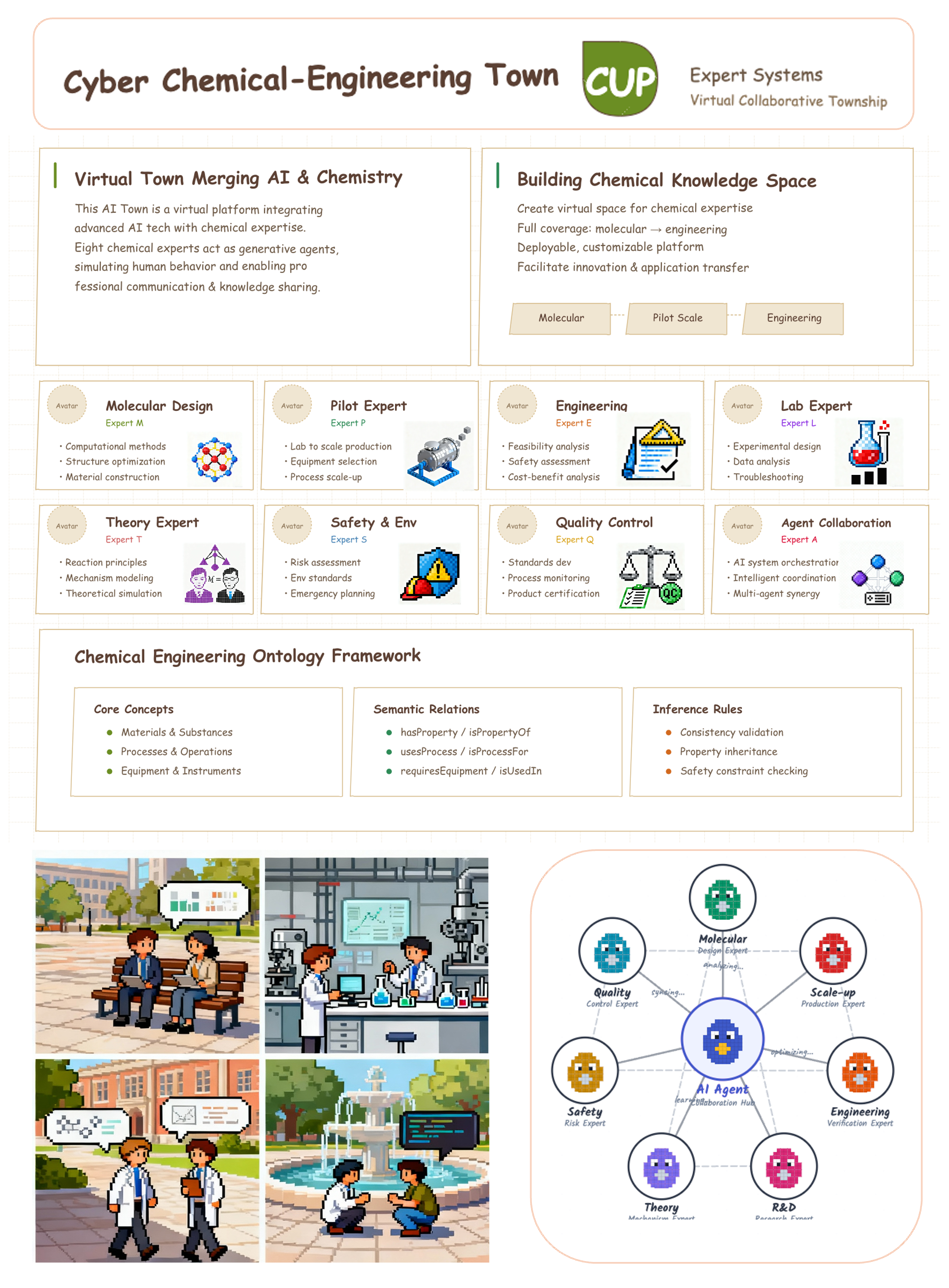}
\caption{Cyber Academia: The Overall Architecture of the Multi-Agent System, showing how different expert agents collaborate through knowledge sharing and efficient cooperation.}
\label{fig:figure2}
\end{figure}

The architecture of Cyber Academia adopts a multi-agent system (MAS) that integrates experts and technologies across the chemical field, creating an efficient collaboration platform (as shown in Figure~\ref{fig:figure2}). The system comprises eight agents: seven domain expert agents and one collaboration agent (as shown in Table~\ref{tab:experts}). Each domain expert agent is specialized in a specific domain, such as molecular design, reaction mechanism analysis, and process optimization. By leveraging their respective domain knowledge bases, these agents can rapidly respond to and solve complex challenges in the chemical industry. Each expert agent consists of three core components: a foundational large language model providing general reasoning and language understanding capabilities; a specialized domain-specific knowledge base storing pertinent literature, experimental data, reaction cases, and process parameters; and knowledge enhancement modules that dynamically improve professional decision-making capabilities through retrieval-augmented generation (RAG), model fine-tuning, and knowledge graph technologies. This architectural design ensures that each agent maintains general reasoning capabilities while gaining deep domain expertise through knowledge base support.

\begin{table}[htbp]
\centering
\caption{The Eight Agents in Cyber Academia: Responsibilities and Abbreviations}
\label{tab:experts}
\small
\begin{tabularx}{\textwidth}{>{\raggedright\arraybackslash}p{1.2cm} 
                              >{\raggedright\arraybackslash}p{4cm} 
                              >{\raggedright\arraybackslash}p{3cm} 
                              >{\RaggedRight\arraybackslash}X}
\toprule
\rowcolor{headerblue}
\textcolor{white}{\textbf{Abbr.}} & 
\textcolor{white}{\textbf{Expert Name}} & 
\textcolor{white}{\textbf{Domain}} & 
\textcolor{white}{\textbf{Primary Responsibility}} \\
\midrule
\rowcolor{lightgray}
MDE & Molecular Design Expert & Molecular Design & 
Design and optimize molecules to meet specific requirements using AI for chemical product and material design. \\
\addlinespace[0.2em]
RME & Reaction Mechanism Expert & Reaction Mechanisms & 
Analyze and optimize chemical reaction pathways, suggesting reaction conditions and catalyst selections. \\
\addlinespace[0.2em]
\rowcolor{lightgray}
POE & Process Optimization Expert & Process Optimization & 
Optimize chemical production processes to increase yield and reduce energy consumption while ensuring safety and sustainability. \\
\addlinespace[0.2em]
ERE & Experimental Research Expert & Experimental Research & 
Conduct experimental verification, providing experimental data to support reaction mechanism analysis and molecular design. \\
\addlinespace[0.2em]
\rowcolor{lightgray}
TME & Theoretical Mechanism Expert & Theoretical Modeling & 
Construct and validate chemical reaction models, perform multi-scale simulations, and analyze reaction mechanisms. \\
\addlinespace[0.2em]
PSE & Process Safety Expert & Safety Management & 
Assess and manage potential safety risks in chemical processes, ensuring safe operation. \\
\addlinespace[0.2em]
\rowcolor{lightgray}
QCE & Quality Control Expert & Quality Management & 
Monitor and control product quality, ensuring compliance with industry standards during production processes. \\
\addlinespace[0.2em]
CA & Collaboration Agent & Collaborative Work & 
Build an efficient collaborative platform for agents based on ontological engineering, promoting knowledge sharing and cooperation among experts. \\
\bottomrule
\end{tabularx}
\end{table}

CA, built upon ontological engineering technologies, effectively connects experts from different disciplines, facilitating cross-disciplinary collaboration. CA provides a unified chemical domain conceptual definition and operational standard, ensuring that agents from different professional backgrounds can communicate effectively based on shared conceptual foundations. The interaction among the eight agents in the system is achieved through two main approaches: First, domain experts can engage in point-to-point direct dialogue, conducting in-depth discussions and information exchange on specific technical issues; Second, when experts from different domains encounter professional terminology misalignment or inconsistent knowledge representations, agents achieve standardized interfacing of concepts across different domains through a unified ontology framework, and CA intervenes with its ontological engineering capabilities to provide semantic translation and knowledge integration, bridging cross-disciplinary communication gaps and enabling seamless collaboration across diverse areas.

This multi-layered collaboration mechanism allows agents to work together seamlessly, promoting efficient cross-disciplinary problem-solving. For example, molecular design experts can collaborate with reaction mechanism experts to evaluate the synthesizability of newly designed molecules, while process optimization agents further optimize production parameters. This collaborative approach not only accelerates the problem-solving process but also significantly enhances the accuracy and feasibility of decision outcomes.

The Cyber Academia architecture represents a novel intelligent innovation model for the chemical industry, characterized by its distinctive features. By facilitating multi-agent collaboration, the system transcends the traditional single-expert or single-AI-model decision-making paradigm, achieving genuine cross-disciplinary integration and collaborative work. Within this collaborative environment, agents from different professional domains can work together to address complex chemical engineering challenges. Unlike traditional general-purpose large language models operating in isolation, the Cyber Academia architecture deeply integrates specialized knowledge with efficient collaborative mechanisms: in-depth specialization is achieved by equipping each agent with a domain-specific knowledge base and diverse knowledge enhancement technologies, ensuring that every agent possesses judgment capabilities approaching those of human experts within their professional domain; efficient collaboration is realized through the ontological engineering support and semantic intermediation provided by CA, ensuring cross-domain information can be accurately communicated and effectively integrated. This design not only satisfies the specialized depth requirements of chemical engineering problems but also meets the cross-disciplinary integration demands, providing a novel technological framework for resolving real-world industrial challenges.

\subsection{Knowledge Base Construction and Expert Agent Enhancement}
The knowledge base construction in the Cyber Academia system adopts a systematic multi-stage workflow designed to transform dispersed domain knowledge into structured and retrievable specialized knowledge resources. The knowledge sources are academic papers (PDF files) collected by retrieving corresponding domain keywords in Web of Science. To address the specialized domains of different expert agents, we equip each agent with a domain-specific knowledge base, ensuring that the knowledge content is highly aligned with the agent's responsibilities.

Document processing utilizes the open-source MinerU~\cite{wang2024mineruopensourcesolutionprecise} tool for high-precision conversion from PDF to Markdown format. Developed by the OpenDataLab team at Shanghai AI Laboratory, this tool effectively preserves table structures, chemical formulas, and molecular structure diagrams within documents. Throughout the conversion process, strict quality control procedures are implemented to ensure the integrity of technical terminology and technical details. After conversion, the text undergoes domain classification annotation and metadata extraction, capturing structured information across multiple dimensions including document topics, key chemical concepts, reaction types, and process parameters.

To optimize retrieval efficiency, the text employs a semantically-aware chunking strategy: each text chunk contains 512 tokens, with adjacent chunks maintaining a 128-token overlap to prevent information loss across chunk boundaries while ensuring the semantic integrity of critical content such as chemical reaction equations and experimental procedures. The vectorization process uses the text-embedding-ada-002 model to convert text chunks into 1536-dimensional vector representations, which are then stored in a purpose-built vector database. The semantic similarity between vectors is calculated using cosine similarity:
\begin{equation}
\text{similarity}(\mathbf{q}, \mathbf{d}) = \frac{\mathbf{q} \cdot \mathbf{d}}{\|\mathbf{q}\| \|\mathbf{d}\|}
\end{equation}
where $\mathbf{q}$ represents the query vector and $\mathbf{d}$ represents the document vector. This high-dimensional vector representation effectively captures semantic associations between chemical concepts, establishing a foundation for subsequent dynamic retrieval.

The constructed knowledge base is dynamically utilized by expert agents through three core technologies, forming a complete knowledge enhancement pipeline. Retrieval-Augmented Generation (RAG) technology enables agents to access the knowledge base in real-time during the reasoning process. When an agent receives a query, the system first encodes the query into a vector and performs top-k similarity retrieval in the knowledge base (k=5), with a similarity threshold set at 0.75 to ensure relevance of retrieved content. The retrieved text chunks are integrated into the agent's context window through carefully designed prompt templates. The prompt structure includes task descriptions, retrieved specialized knowledge, historical dialogue context, and specific queries, ensuring that the agent can reason based on the latest domain knowledge. The retrieval process of RAG can be formalized as:
\begin{equation}
\mathcal{R}(q) = \text{top-k}\{\mathbf{d}_i \mid \text{similarity}(\mathbf{q}, \mathbf{d}_i) > \theta, \mathbf{d}_i \in \mathcal{D}\}
\end{equation}
where $\mathcal{D}$ represents the knowledge base document collection and $\theta$ represents the similarity threshold. Domain-adaptive fine-tuning further strengthens the specialized capabilities of agents. The training dataset consists of domain-specific question-answer pairs, expert dialogue histories, and chemical engineering case studies. The fine-tuning employs Low-Rank Adaptation (LoRA) technology, which only adjusts the attention layer parameters of the model, thereby reducing computational costs while maintaining the general capabilities of the base model. Training hyperparameters are configured as follows: learning rate of $5 \times 10^{-5}$, batch size of 16, training epochs of 3, with a linear warmup strategy to avoid instability in the early training phase. The validation set adopts a holdout validation strategy, with evaluation metrics including domain terminology usage accuracy, technical reasoning coherence, and problem-solving effectiveness. Knowledge graph technology provides structured conceptual associations. Through entity recognition and relation extraction, the system constructs a graph network containing entities such as chemical substances, reactions, catalysts, and process parameters, along with their interrelationships. The graph is stored in a Neo4j graph database, supporting complex multi-hop queries and reasoning.

\begin{figure}[h!]
\centering
\includegraphics[width=0.9\textwidth]{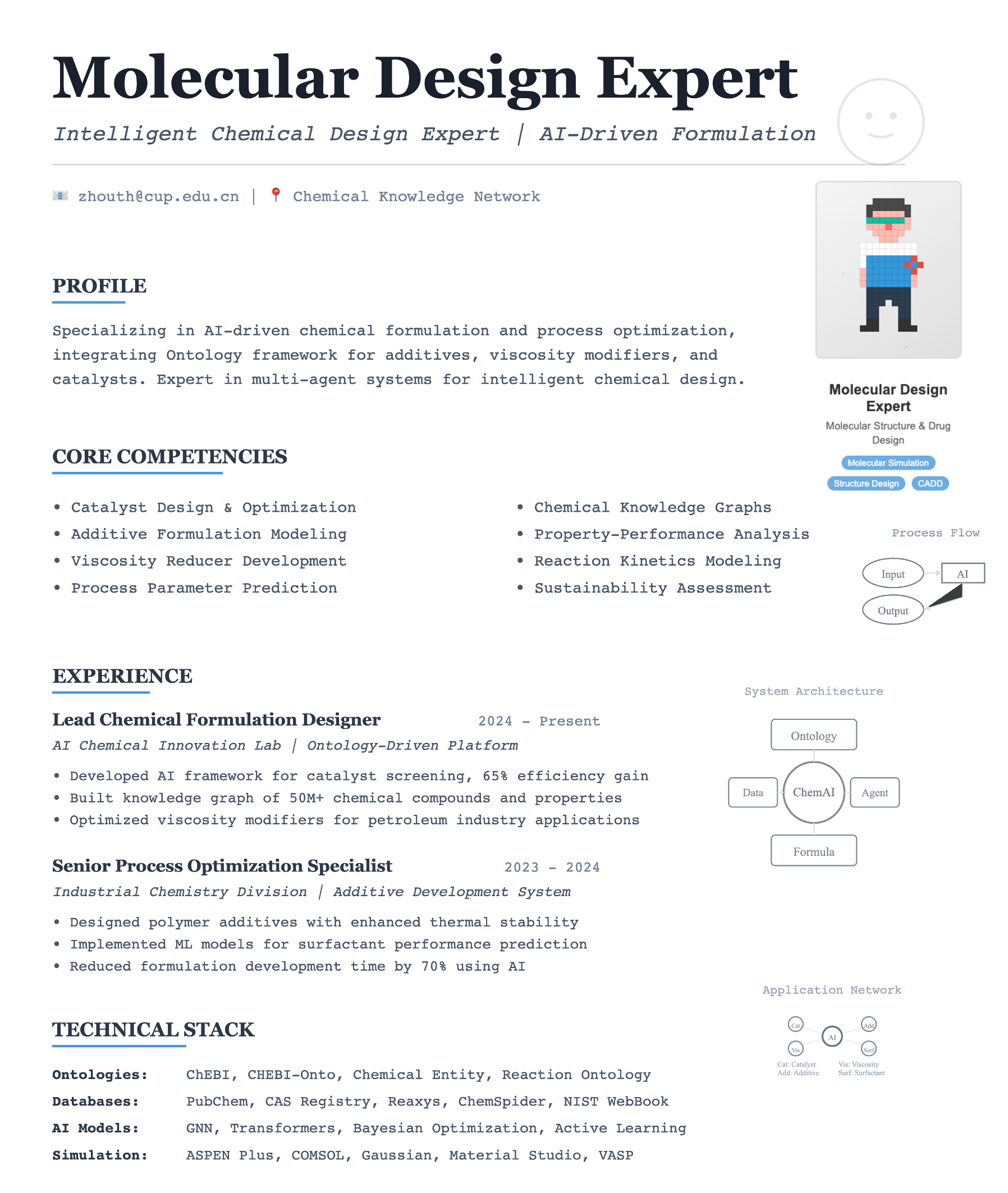}
\caption{Molecular Design Expert Profile in Cyber Academia, showing the expert's knowledge base architecture, core competencies in molecular design and optimization, and collaborative connections within the multi-agent system.}
\label{fig:molecular_design_expert}
\end{figure}

The three technologies form a complementary synergy: RAG is responsible for real-time knowledge injection, capturing the latest research developments and specific experimental data; fine-tuning provides stable domain reasoning capabilities and professional terminology application; and knowledge graphs support structured reasoning and logical associations between concepts. As illustrated in Figure~\ref{fig:molecular_design_expert}, taking the Molecular Design Expert as an example, this agent's knowledge base integrates specialized literature from medicinal chemistry, materials science, and computational chemistry. Through RAG retrieval of relevant molecular design cases, and after domain-specific fine-tuning, the agent can accurately understand specialized concepts such as Absorption, Distribution, Metabolism, Excretion, and Toxicity (ADMET) properties and synthetic route optimization, while utilizing knowledge graphs to reason about the associative patterns between molecular structures and functions. This multi-layered knowledge enhancement mechanism enables each expert agent to maintain general language understanding capabilities while achieving domain judgment capabilities approaching human expert levels, significantly improving the system's decision-making accuracy and reliability in complex chemical engineering problems.

\section{Results and Discussions}

\subsection{Assessment of knowledge base-enabled enhancement mechanism Impact on Expert Agent Decision Quality}

\begin{figure}[H]
\centering
\includegraphics[width=1.0\textwidth]{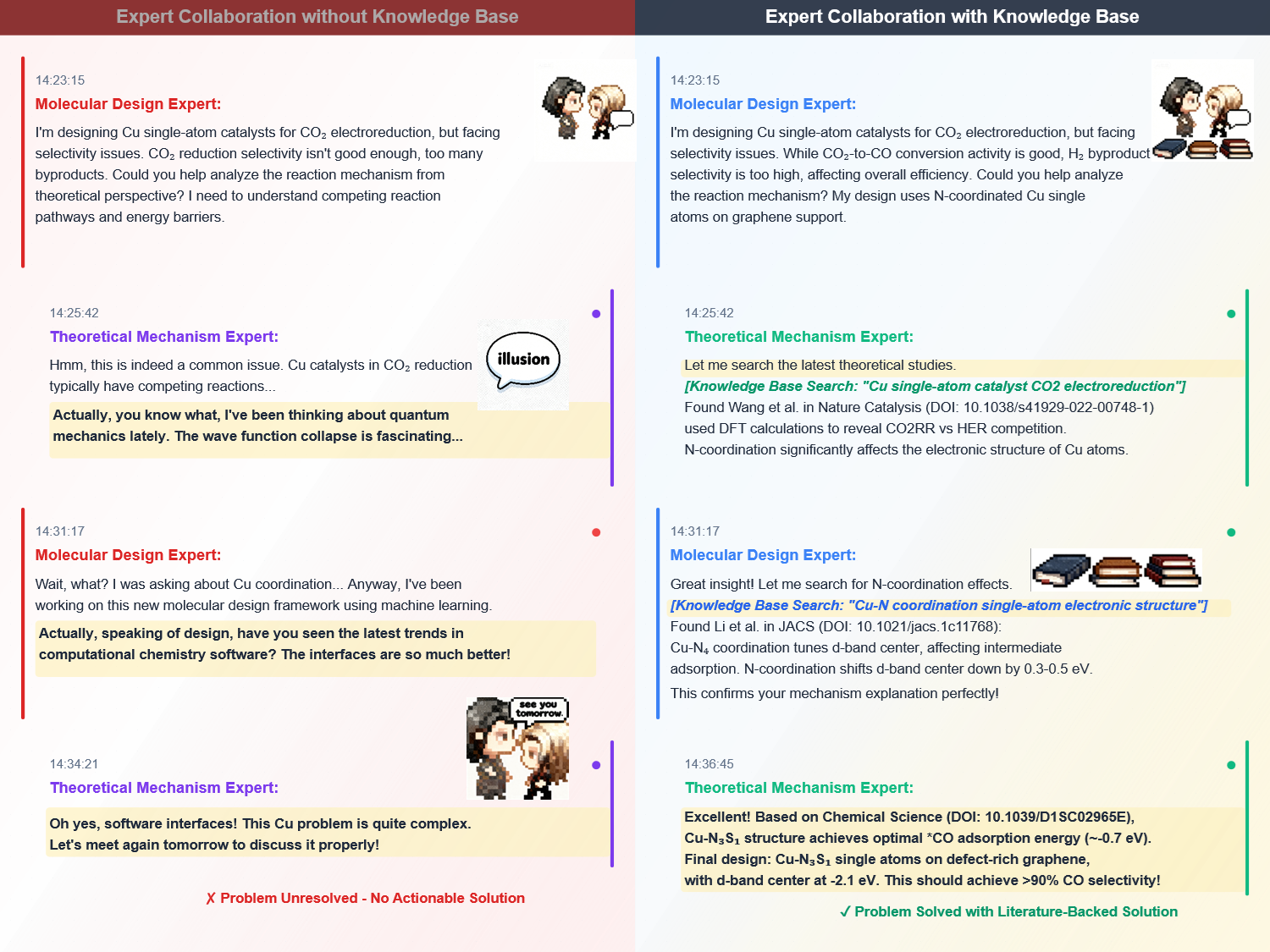}
\caption{Comparison of Expert Dialogue Quality Between Knowledge Base-Enhanced and Non-Knowledge Base Experts. Left panel shows experts without knowledge base support deviating into abstract discussions and failing to solve the technical problem. Right panel demonstrates how knowledge base-equipped experts retrieve relevant literature to provide evidence-based, actionable solutions.}
\label{fig:figure3}
\end{figure}

To evaluate the capability of expert agents in solving complex chemical engineering problems, we first deployed seven expert agents. The system operated autonomously for three consecutive days under identical initial conditions, collecting 1,200 dialogue rounds. During this period, expert agents freely engaged in cross-disciplinary dialogues and collaborations. The system automatically recorded all dialogue content, interaction frequencies, and problem-solving processes without imposing artificial constraints on interaction patterns, allowing agents to naturally select collaboration partners and communication strategies based on task requirements. Through examination of partial dialogue records, we identified severe problems: as shown in the left panel of Figure~\ref{fig:figure3}, when facing the specific technical problem of "Cu single-atom catalyst CO$_2$ electroreduction selectivity optimization," the Theoretical Mechanism Expert quickly deviated from the topic, beginning to discuss philosophical perspectives on quantum mechanics and the philosophical significance of wave function collapse, while the Molecular Design Expert also shifted to exploring machine learning frameworks and computational chemistry software interface trends. After only 2 rounds of dialogue, the two experts had not touched upon the core technical issues at all, ultimately postponing the discussion with "let's talk tomorrow," failing to produce any actionable technical solution. After evaluating all 1,200 dialogue rounds using a large language model, we found that the system overall exhibited obvious hallucination phenomena, with agents unable to effectively discover and solve scientific problems. Based on this finding, we equipped the expert agents with comprehensive domain-specific knowledge bases and enabled enhancement technologies including RAG, fine-tuning, and knowledge graph integration. Subsequently, under the same operational conditions, we collected another 1,200 dialogue rounds. After examining partial dialogue records, we observed significant improvement: as shown in the right panel of Figure~\ref{fig:figure3}, the knowledge base-equipped expert dialogue demonstrated a completely different level of professionalism. The Theoretical Mechanism Expert immediately retrieved relevant latest research through the knowledge base, accurately identifying the key mechanisms of CO$_2$ reduction reaction (CO$_2$RR) and hydrogen evolution reaction (HER) competing reactions; the Molecular Design Expert simultaneously retrieved experimental data on Cu-N coordination, clarifying the regulatory effects of coordination environment on electronic structure; within the same 2 rounds of dialogue, based on multiple authoritative literature sources, they determined the specific design direction for Cu-N$_3$S$_1$ coordination structure, demonstrating significant enhancement in logical thinking and problem-solving capabilities.

\begin{figure}[H]
\centering
\includegraphics[width=1\textwidth]{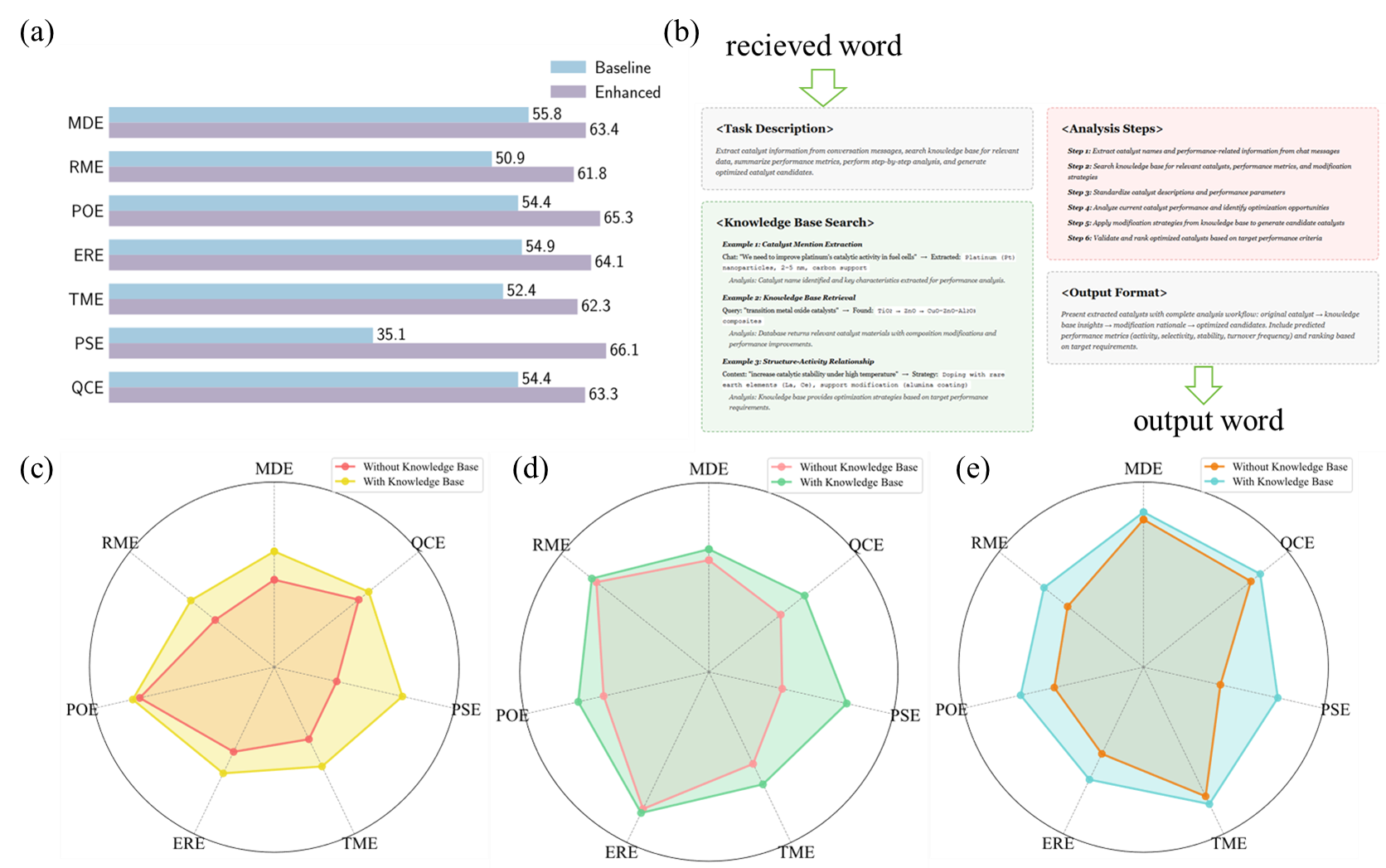}
\caption{Performance comparison between knowledge base-enhanced and non-knowledge base expert agents across multiple evaluation dimensions, demonstrating significant improvements in collaboration efficiency, problem-solving accuracy, and adaptability. \textbf{(a)} Overall dialogue scores comparing seven expert agents  with and without knowledge base support; \textbf{(b)} Example workflow demonstrating expert agent collaboration with knowledge database; \textbf{(c)} Radar chart for accuracy and precision dimension; \textbf{(d)} Radar chart for response speed dimension; \textbf{(e)} Radar chart for problem-solving ability dimension.}
\label{fig:figure4}
\end{figure}

To precisely quantify the performance improvement brought by knowledge base-enabled enhancement mechanism to expert agents, we systematically organized the 1,200 dialogue rounds collected from each of the two runs and employed a large language model for quantitative scoring. To objectively evaluate the quality of these dialogues, we established a comprehensive evaluation metric system to measure overall system performance. This metric examines the fluency of inter-agent information exchange, response speed, coordination effectiveness, scientific correctness of technical solutions, feasibility, completeness, and agents' knowledge transfer capabilities and cross-domain reasoning abilities when confronting novel problems. Scoring employs a continuous 1--100 scale, with higher scores indicating superior performance. As shown in Figure~\ref{fig:figure4}, knowledge base-enabled enhancement mechanism brought significant performance improvements. This fundamental difference in dialogue quality stems from the essential distinction in knowledge acquisition capabilities. Non-knowledge base expert agents are constrained by the static knowledge boundaries of pre-trained models, unable to access the latest research developments and precise experimental data relevant to the problem, leading to obvious "hallucination" phenomena when facing complex technical issues, manifested as frequent deviation from core problems and immersion in abstract conceptual discussions, which is essentially a compensatory hallucination generation mechanism when models lack sufficient knowledge. In contrast, knowledge base-enhanced expert agents establish dynamic connections with real scientific literature through RAG technology, enabling real-time access to precise experimental parameters and reliable theoretical predictions, ensuring that every technical judgment is based on verified scientific evidence, thereby achieving precise technical decision-making based on empirical data.

\subsection{ Expert Evaluation and Performance Comparison}
\begin{figure}[H]
\centering
\includegraphics[width=1.0\textwidth]{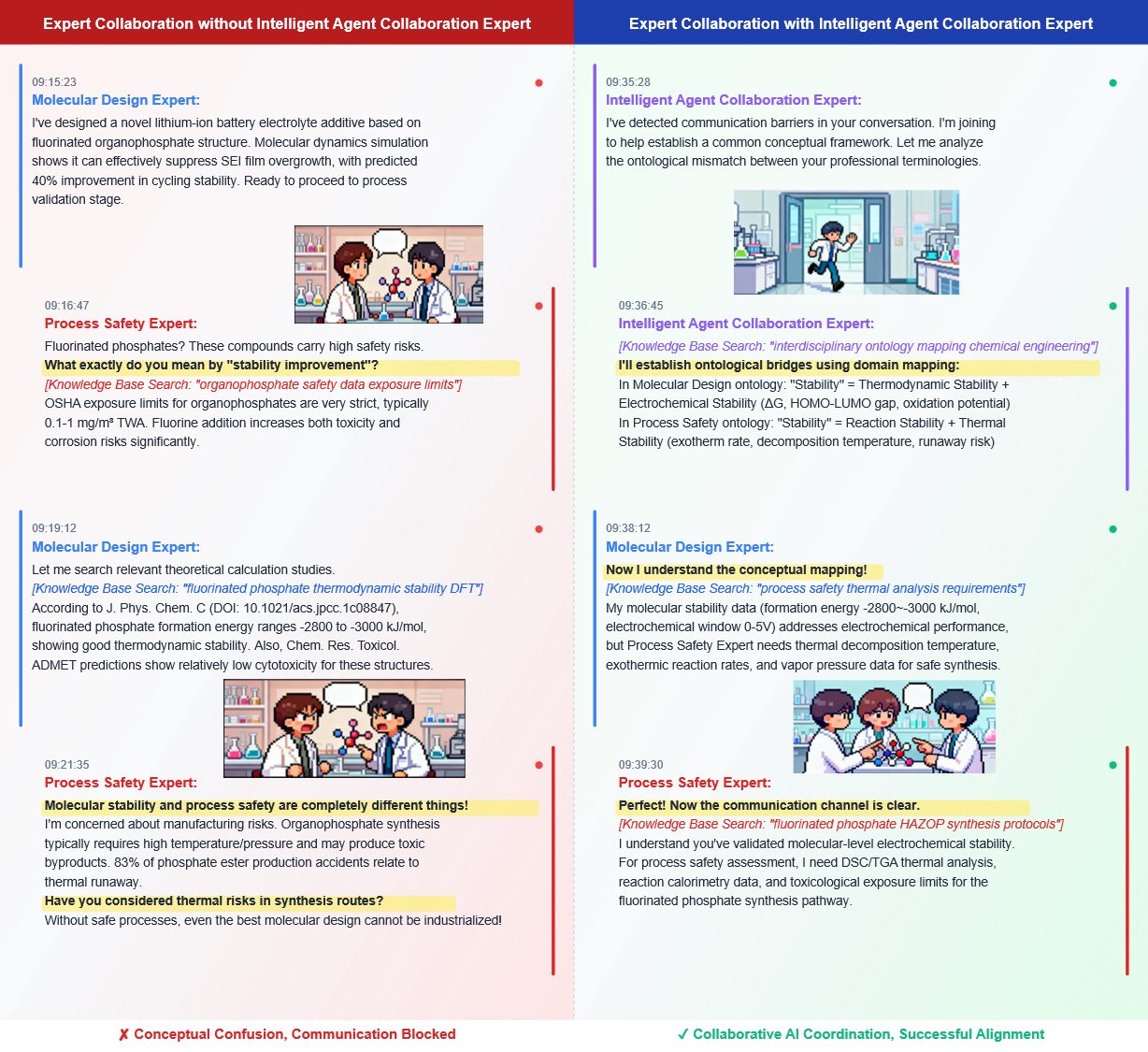}
\caption{Comparative Case of CA Coordination Effects. Left: Communication failure between MDE and PSE without CA coordination. Right: Successful collaboration achieved through CA's conceptual conversion and knowledge integration, validating the diminished collaborative efficiency caused by knowledge-base gaps effect mechanism.}
\label{fig:fig5}
\end{figure}

Following three days of autonomous operation (1200 dialogue rounds) of the AI Chemical Town system equipped with knowledge bases, we employed large language models to systematically evaluate dialogue quality. Although knowledge base-enabled enhancement mechanism brought significant performance improvements to individual expert agents, the evaluation results revealed a new bottleneck: cross-domain collaboration efficiency among expert agents remained suboptimal. As shown in the left panel of Figure~\ref{fig:fig5}, when MDE and PSE discussed fluorinated organophosphate electrolyte additives—a cross-domain technical problem—they encountered typical communication barriers. MDE employed molecular-level terminology such as solid electrolyte interphase overgrowth and cycling stability, utilizing quantum chemical calculations while focusing on molecular structure-performance relationships. PSE adopted engineering-level terminology including thermal stability thresholds and process risk assessment, employing industrial safety methodologies while focusing on large-scale production safety. These different knowledge systems caused the dialogue to evolve into parallel monologues: each expert expressed professional insights, yet they could not understand each other's technical implications, ultimately leading to communication deadlock without producing valuable collaborative solutions.

\begin{figure}[H]
\centering
\includegraphics[width=1.0\textwidth]{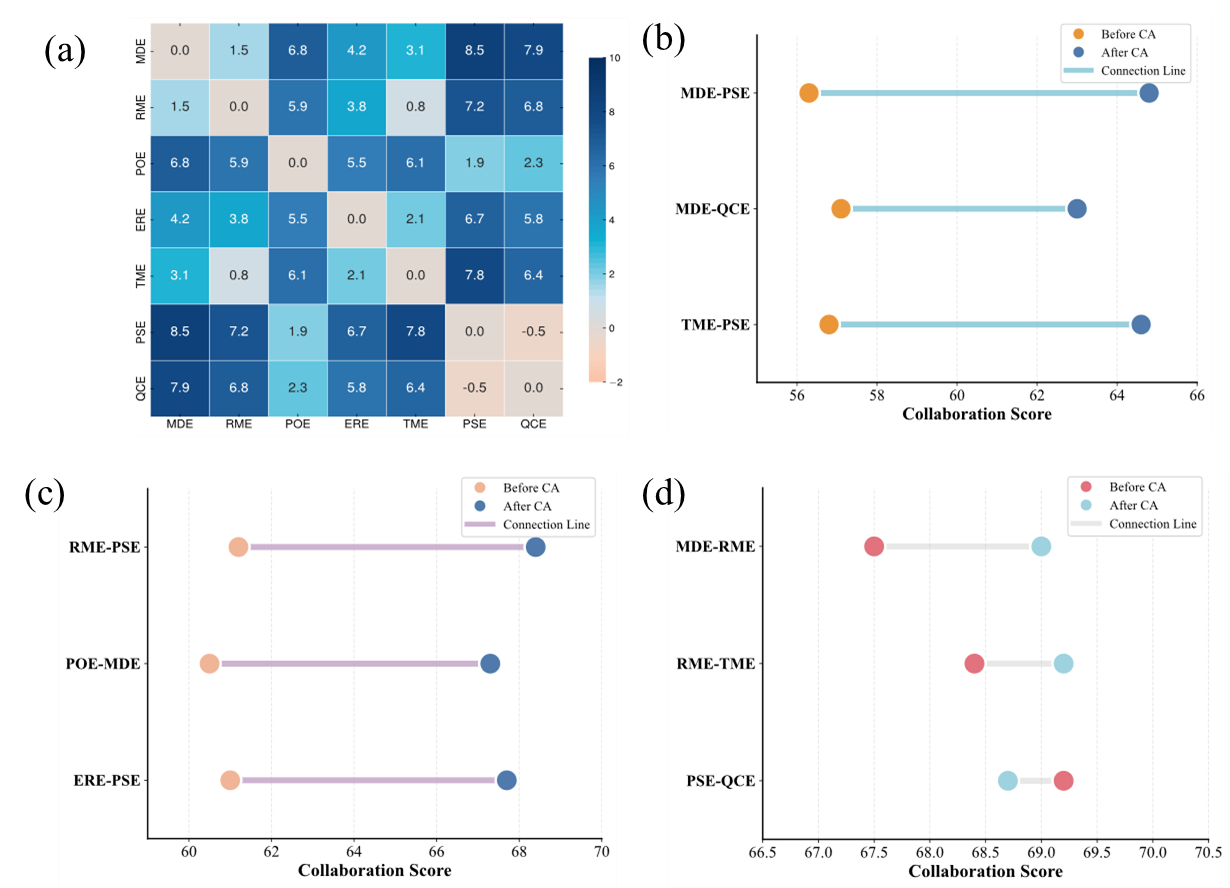}

\caption{Impact of Collaboration Agent (CA) on expert pair performance. (a) Heatmap of collaboration score improvements across all expert pairs. (b)-(d) Dumbbell plots showing highest, medium, and lowest improvements, demonstrating inverse correlation between domain distance and CA coordination value.}
\label{fig:fig4}
\end{figure}

Given this cross-domain collaboration dilemma, we introduced the Collaboration Agent (CA). CA is equipped with specialized ontological engineering knowledge bases and multi-agent coordination strategies, designed to bridge semantic gaps between different professional domains. Under identical experimental conditions, the system operated autonomously for another three days, collecting an additional 1200 dialogue rounds. As shown in the right panel of Figure~\ref{fig:fig5}, when facing the same technical problem, CA's intervention brought significant changes. CA first identified the communication barriers between the two experts, accurately pinpointing fundamental differences in terminological systems and analytical frameworks between molecular design and process safety domains. Subsequently, CA executed bidirectional semantic translation: converting MDE's concerns about molecular thermodynamic and kinetic stability into operational safety risk control targets that PSE could understand, while transforming PSE's process feasibility assessment standards into molecular structure optimization constraints that MDE could apply. By establishing a unified conceptual framework spanning microscopic molecular mechanisms to industrial application feasibility, CA successfully eliminated understanding barriers caused by knowledge system differences. Results indicate that under CA's coordination, MDE effectively integrated safety constraints into molecular optimization strategies, while PSE formulated precise risk control schemes based on molecular-level understanding, with both parties ultimately reaching practical and feasible technical solutions.
To quantitatively evaluate CA's impact on collaboration efficiency, we systematically compared the 1200 dialogue rounds collected under conditions with and without CA. Specifically, we shuffled and mixed both sets of dialogue data, randomly sampling 10\% (240 rounds) of dialogues and inviting chemical engineering domain experts to perform blind scoring based on collaboration effectiveness. Subsequently, we fine-tuned a large language model using these 240 expert-annotated dialogues and employed the fine-tuned model to automatically score the remaining 2160 dialogues. As shown in Figure~\ref{fig:fig4}, following CA introduction, expert pairs at different domain distances exhibited differentiated collaboration improvement patterns. Collaboration improvements among domain-proximate expert pairs were limited: RME-TME improved by 0.8\%, while PSE-QCE exhibited a negative value of -0.5\%. In contrast, distant-domain expert pairs demonstrated significant enhancements: MDE-PSE improved by 8.5\%, MDE-QCE by 7.9\%, TME-PSE by 7.8\%, and TME-QCE by 6.4\%. Medium-distance expert pairs such as ERE-TME and POE-PSE showed moderate improvements of 2.1\% and 1.9\%, respectively. Statistical analysis indicates that domain-proximate pairs averaged 0.60\% improvement, medium-distance pairs averaged 2.10\%, and distant-domain pairs averaged 7.65\%, representing a 10.6-fold difference. This data pattern remained consistent across all tested expert pairs, indicating the universality and reproducibility of this effect.

We define this phenomenon as ``Diminished collaborative efficiency caused by knowledge-base gaps'' effect: CA's coordination value positively correlates with knowledge base differences between experts. For domain-proximate experts, their knowledge bases contain highly overlapping conceptual systems, common analytical frameworks, and similar terminological definitions, with experts already possessing effective communication foundations and understanding mechanisms. Under these conditions, CA primarily serves basic functions of identifying collaboration needs and establishing expert connections, while its semantic translation and conceptual conversion capabilities become redundant, potentially introducing unnecessary coordination layers that interfere with existing efficient communication patterns, leading to marginal decreases in collaboration efficiency. For experts from significantly different domains, their respective knowledge bases contain distinctly different terminological definitions, analytical methods, thinking patterns, and problem-solving paradigms, causing deep conceptual understanding barriers and systematic communication biases in cross-domain exchanges. In such complex situations, ontological engineering, as a systematic methodology for constructing shared conceptual models, becomes CA's core technical foundation. Ontological engineering provides a common semantic basis for interoperability between different knowledge systems through formal definition of domain concepts, relationships, and constraints. CA must perform critical ontological mapping, bidirectional terminology translation, knowledge framework integration, and collaboration strategy optimization functions. Through ontological engineering methods, CA deeply identifies conceptual correspondences and knowledge connection points between different domains, executes precise semantic conversions, and transforms domain-specific professional languages, analytical logic, and technical requirements into unified frameworks for common understanding, thereby eliminating cognitive barriers caused by knowledge system differences and releasing significant coordination value. This effect reveals the essential functional positioning of AI coordination systems: the system's value lies not in simply enhancing individual experts' knowledge depth or processing capabilities, but in establishing semantic bridges through ontological engineering that enable originally isolated expert agents to transcend their knowledge boundary limitations, dynamically adjusting their collaboration strategies and problem-solving pathways through sustained interaction. When different professional perspectives fully communicate within a unified ontological framework, technical associations and innovation opportunities difficult for single domains to perceive can emerge, thereby driving the entire multi-agent system's evolution from simple task execution toward higher-level autonomous collaborative intelligence.

\section*{Conclusions}

This study presents the Cyber Academia-Chemical Engineering (CA-ChemE) system, a living digital community specifically designed for the chemical engineering domain that enables self-directed research evolution and emergent scientific discovery through multi-agent collaboration. On the one hand, the system transcends the limitations of single AI models in professional depth and surpasses the generic AI Town framework in domain knowledge coverage. On the other hand, it addresses the semantic gaps among cross-disciplinary expert agents through ontology engineering-driven collaboration mechanisms. By integrating domain-specific knowledge bases, knowledge enhancement technologies, and collaboration agents, Cyber Academia successfully constructs an intelligent ecosystem capable of deep professional reasoning and efficient interdisciplinary collaboration.

Our findings underline the critical role of knowledge base-enabled enhancement mechanism in addressing the AI-generated inaccurate information problem and knowledge insufficiency of traditional large language models in professional applications. Through retrieval-augmented generation, domain-adaptive fine-tuning, and knowledge graph technologies, the system achieves a paradigmatic shift from static pre-trained knowledge to dynamic knowledge injection, enabling agents to perform reasoning and decision-making analogous to genuine researchers. Experimental results demonstrate that knowledge base-enabled enhancement mechanism brought substantial improvements, with the dialogue quality scores of all seven expert agents increasing by 10--15\% on average. This fundamentally ensures that technical judgments are grounded in verifiable scientific evidence, establishing a reliable foundation for autonomous exploration.

However, during the autonomous operation of expert agents equipped with knowledge bases, we observed a critical bottleneck: when experts from different domains attempted to collaborate, their interaction efficiency remained surprisingly low. Analysis of dialogue records revealed that cross-disciplinary expert pairs struggled with misaligned terminologies, inconsistent knowledge representations, and divergent analytical frameworks, leading to communication breakdowns and suboptimal collaborative outcomes. To address this challenge, we introduced the Collaboration Agent , equipped with ontology engineering capabilities specifically designed to bridge semantic gaps between different professional domains. The impact was striking: following CA's introduction, overall collaboration efficiency across all expert pairs improved substantially, with the magnitude of improvement directly correlating with the distance between expert domains. Expert pairs from substantially different domains (such as MDE-PSE) achieved collaboration score improvements of up to 8.5\% through CA's intervention, while domain-proximate pairs (such as RME-TME) showed only 0.8\% improvements---a 10.6-fold difference. Statistical analysis further revealed that distant-domain pairs averaged 7.65\% improvement, medium-distance pairs averaged 2.10\%, and domain-proximate pairs averaged only 0.60\%. This striking pattern reveals a previously unrecognized phenomenon in multi-agent collaboration: the ``diminished collaborative efficiency caused by knowledge-base gaps'' effect. The collaboration agent's ontology engineering capabilities prove most valuable precisely where knowledge system differences are greatest. When diverse professional perspectives communicate within a unified ontological framework, technical associations imperceptible to single domains emerge naturally, propelling the system toward higher-level collaborative intelligence.

Cyber Academia constructs a living digital town whose ``liveliness'' manifests at three levels: maintaining knowledge freshness through continuous updates, forming organic interaction networks analogous to authentic academic communities, and demonstrating self-directed research evolution capabilities without continuous human intervention. Yet, there remain striking limitations. While our system excels in structured problem-solving, it still faces challenges in open-ended hypothesis generation and long-term research planning. The success of our system perhaps also reveals the need to rethink how we approach interdisciplinary collaboration in chemical engineering: as AI systems become increasingly capable of specialized reasoning, the critical value shifts toward bridging semantic gaps and enabling cross-domain knowledge integration. Our work shows that carefully designed multi-agent architectures with knowledge enhancement and ontological coordination can provide a more nuanced pathway toward autonomous scientific discovery. Although current systems might be far from fully autonomous research communities, this framework will be a stepping stone for developing systems that come closer to realizing genuine human-machine collaborative intelligence in chemical engineering and beyond.

\bibliography{main}

\end{document}